\documentclass[11pt,a4paper]{article}
\usepackage[hyperref]{naaclhlt2018}
\usepackage{times}
\usepackage{latexsym}
\usepackage{graphicx}
\usepackage{url}
\usepackage[utf8]{inputenc}

\aclfinalcopy 

\title{Tracking Typological Traits of Uralic Languages\\in Distributed Language Representations}

\author{Johannes Bjerva\\
Department of Computer Science\\
University of Copenhagen\\
Denmark\\
{\tt bjerva@di.ku.dk} \\\And
Isabelle Augenstein\\
Department of Computer Science\\
University of Copenhagen\\
Denmark\\
{\tt augenstein@di.ku.dk}
\\}

\date{}

\begin{document}
\maketitle
\begin{abstract}
Although linguistic typology has a long history, computational approaches have only recently gained popularity.
The use of distributed representations in computational linguistics has also become increasingly popular.
A recent development is to learn distributed representations of language, such that typologically similar languages are spatially close to one another.
Although empirical successes have been shown for such language representations, they have not been subjected to much typological probing.
In this paper, we first look at whether this type of language representations are empirically useful for model transfer between Uralic languages in deep neural networks.
We then investigate which typological features are encoded in these representations by attempting to predict features in the \textit{World Atlas of Language Structures}, at various stages of fine-tuning of the representations.
We focus on Uralic languages, and find that some typological traits can be automatically inferred with accuracies well above a strong baseline.

\begin{center}
\section*{\centering Tiivistelmä}
\end{center}
Vaikka kielitypologialla on pitkä historia, laskentamenetelmät ovat vasta viime aikoina saavuttaneet suosiota.
Myös hajautettujen esitysten käyttö laskennallisessa kielitieteessä on tullut suositummaksi.
Viimeaikainen kehitys on hajautetun kieliedustuksen oppiminen, kuten että samanlaiset kielet ovat lähellä toisiaan.
Vaikka empiirisiä tuloksia onkin saavutettu, ei niille ole tehty paljoakaan typologista tutkimusta.
Tässä artikkelissa tutkitaan ensin, ovatko tämänlaiset kieliedustukset empiirisesti käyttökelpoisia, kun kyseessä on uralilaisten kielten "model transfer" syvissä neuroverkoissa.
Tutkimme myös, mitä typologisia piirteitä voimme löytää kieliedustuksissa, yrittämällä ennustaa ominaisuuksia jotka saamme \textit{World Atlas of Language Structures}:n kautta.
Keskitymme uralilaisiin kieliin ja löydämme, että jotkin typologiset piirteet voidaan automaattisesti päätellä selvästi vahvan perustason yläpuolelle.
\end{abstract}


\section{Introduction}
For more than two and a half centuries, linguistic typologists have studied languages with respect to their structural and functional properties, thereby implicitly classifying languages as being more or less similar to one another, by virtue of such properties \citep{haspelmath:2001,velupillai:2012}.
Although typology has a long history \citep{herder:1772,gabelentz:1891,greenberg:1960,greenberg:1974,dahl:1985,comrie:1989,haspelmath:2001,croft:2002}, computational approaches have only recently gained popularity \citep{dunn:2011,walchli:2014,ostling:2015,bjerva:2016:cl4lc,deri:2016,cotterell:2017,peters:2017,asgari:typology,malaviya:2017}.
One part of traditional typological research can be seen as assigning sparse explicit feature vectors to languages, for instance manually encoded in databases such as the World Atlas of Language Structures (WALS, \citealp{wals}).
A recent development which can be seen as analogous to this, is the process of learning distributed language representations in the form of dense real-valued vectors, often referred to as \textit{language embeddings} \citep{tsvetkov:2016,ostling_tiedemann:2017,malaviya:2017}.
These language embeddings encode typological properties of language, reminiscent of the sparse features in WALS, or even of parameters in Chomsky's Principles and Parameters framework \citep{chomsky:1993,chomskylesnik:1993,chomsky:2014}.

In this paper, we investigate the usefulness of explicitly modelling similarities between languages in deep neural networks using language embeddings.
To do so, we view NLP tasks for multiple Uralic languages as different aspects of the same problem and model them in one model using multilingual transfer in a multi-task learning model.
Multilingual models frequently follow a hard parameter sharing regime, where all hidden layers of a neural network are shared between languages, with the language either being implicitly coded in the input string \citep{googlenmt}, given as a language ID in a one-hot encoding \citep{onemodel}, or as a language embedding \citep{ostling_tiedemann:2017}. 
In this paper, we both explore multilingual modelling of Uralic languages, and probe the language embeddings obtained from such modelling in order to gain novel insights about typological traits of Uralic languages.
We aim to answer the following three research questions (\textbf{RQ}s).

\begin{enumerate}
    \item [{\bf RQ~1}] To what extent is model transfer between Uralic languages for PoS tagging mutually beneficial?
    \item [{\bf RQ~2}] Are distributed language representations useful for model transfer between Uralic languages?
    \item [{\bf RQ~3}] Can we observe any explicit typological properties encoded in these distributed language representations when considering Uralic languages?
\end{enumerate}

\section{Data}

\subsection{Distributed language representations}
There are several methods for obtaining distributed language representations by training a recurrent neural language model \citep{rnnlm} simultaneously for different languages \citep{tsvetkov:2016,ostling_tiedemann:2017}.
In these recurrent multilingual language models with long short-term memory cells (LSTM, \citealp{lstm}), languages are embedded into a $n$-dimensional space.
In order for multilingual parameter sharing to be successful in this setting, the neural network is encouraged to use the language embeddings to encode features of language.
Other work has explored learning language embeddings in the context of neural machine translation  \citep{malaviya:2017}.
In this work, we explore the embeddings trained by \citet{ostling_tiedemann:2017}, both in their original state, and by further tuning them for PoS tagging. 

\subsection{Part-of-speech tagging}
We use PoS annotations from version 2 of the Universal Dependencies \citep{nivre:2016}.
We focus on the four Uralic languages present in the UD, namely
Finnish (based on the Turku Dependency Treebank, \citealp{ud:finnish}), Estonian \citep{ud:estonian}, Hungarian (based on the Hungarian Dependency Treebank, \citealp{hundep}), and North Sámi \citep{ud:north_sami}.
As we are mainly interested in observing the language embeddings, we down-sample all training sets to 1500 sentences (approximate number of sentences in the Hungarian data), so as to minimise any size-based effects.

\subsection{Typological data}
In the experiments for RQ3, we attempt to predict typological features.
We extract the features we aim to predict from WALS \citep{wals}.
We consider features which are encoded for all four Uralic languages in our sample.

\section{Method and experiments}

We approach the task of PoS tagging using a fairly standard bi-directional LSTM architecture, based on \citet{plank:2016}.
The system is implemented using DyNet \citep{dynet}.
We train using the Adam optimisation algorithm \citep{adam} over a maximum of 10 epochs, using early stopping.
We make two modifications to the bi-LSTM architecture of  \citet{plank:2016}.
First of all, we do not use any atomic embedded word representations, but rather use only character-based word representations.
This choice was made so as to encourage the model not to rely on language-specific vocabulary.
Additionally, we concatenate a pre-trained language embedding to each word representation.
That is to say, in the original bi-LSTM formulation of \citet{plank:2016}, each word $w$ is represented as $\vec{w}+LSTM_{c}(w)$, where $\vec{w}$ is an embedded word representation, and $LSTM_{c}(w)$ is the final states of a character bi-LSTM running over the characters in a word.
In our formulation, each word $w$ in language $l$ is represented as $LSTM_{c}(w)+\vec{l}$, where $LSTM_{c}(w)$ is defined as before, and $\vec{l}$ is an embedded language representation.
We use a two-layer deep bi-LSTM, with 100 units in each layer.
The character embeddings used also have 100 dimensions.
We update the language representations, $\vec{l}$, during training.
The language representations are 64-dimensional, and are initialised using the language embeddings from \citet{ostling_tiedemann:2017}.
All PoS tagging results reported are the average of five runs, each with different initialisation seeds, so as to minimise random effects in our results.

\subsection{Model transfer between Uralic languages}

\begin{figure*}[htbp]
	\centering
	\includegraphics[width=0.7\textwidth]{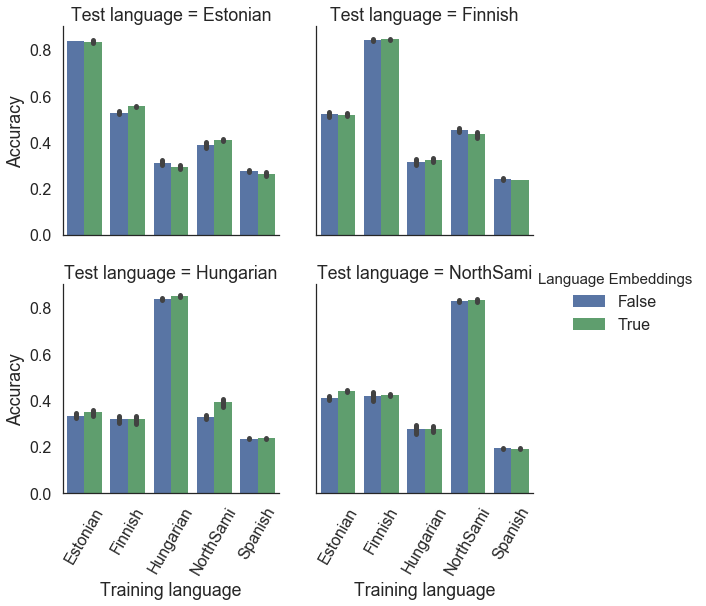}
    \caption{\label{fig:monolingual}Monolingual PoS training. The x-axes denote the training languages, and the y-axes denote the PoS tagging accuracy on the test language at hand.}
\end{figure*}

The aim of these experiments is to provide insight into \textbf{RQ 1} and \textbf{RQ 2}.
We first train a monolingual model for each of the four Uralic languages.
This model is then evaluated on all four languages, to investigate how successful model transfer between pairs of languages is.
Results are shown in Figure~\ref{fig:monolingual}.
Comparing results within each language shows that transfer between Finnish and Estonian is the most successful.
This can be expected considering that these are the two most closely related languages in the sample, as both are Finnic languages.
Model transfer both to and from the more distantly related languages Hungarian and North Sámi is less successful.
There is little-to-no difference in this monolingual condition with respect to whether or not language embeddings are used.
As a baseline, we include transfer results when training on Spanish, which we consider a proxy of a distantly related languages.
Transferring from Spanish is significantly worse ($p<0.05$) than transferring from a Uralic language in all settings.

\begin{figure*}[htbp]
	\centering
	\includegraphics[width=0.7\textwidth]{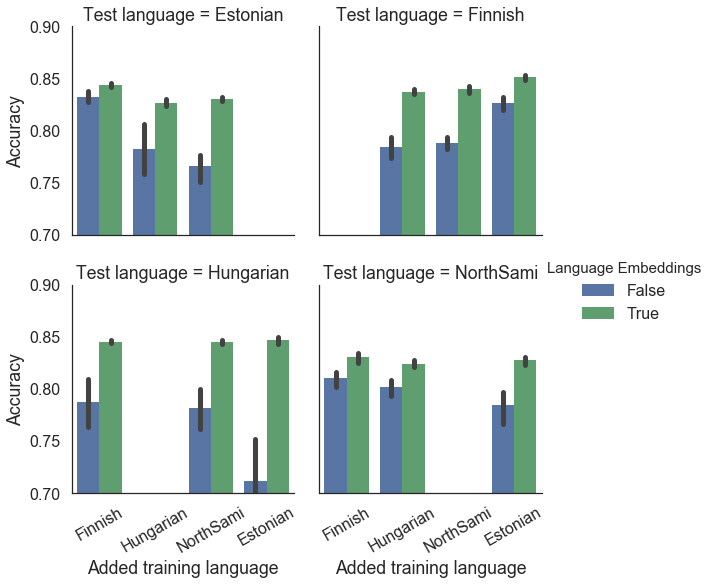}
    \caption{\label{fig:bilingual}Bilingual PoS training. The x-axes denote the added training languages (in addition to the target language), and the y-axes denote the PoS tagging accuracy on the test language at hand.}
\end{figure*}

Next, we train a bilingual model for each Uralic language.
Each model is trained on the target language in addition to one other Uralic language.
Results are shown in Figure~\ref{fig:bilingual}.
Again, transfer between the two Finnic languages is the most successful.
Here we can also observe a strong effect of whether or not language embeddings are incorporated in the neural architecture.
Including language embeddings allows for both of the Finnic languages to benefit significantly ($p<0.05$) from the transfer setting, as compared to the monolingual setting.
No significant differences are observed for other language pairs.


\subsection{Predicting typological features with language embeddings}
Having observed that language embeddings are beneficial for model transfer between Uralic languages, we turn to the typological experiments probing these embeddings.
The aim of these experiments is to provide insight into \textbf{RQ 3}.
We investigate typological features from WALS \citep{wals},  focussing on those which have been encoded for the languages included in the UD.

We first train the same neural network architecture as for the previous experiments on all languages in UD version 2.
Observing the language embeddings from various epochs of training permits tracking the typological traits encoded in the distributed language representations as they are fine-tuned.
In order to answer the research question, we train a simple linear classifier to predict typological traits based on the embeddings.
Concretely, we train a logistic regression model, which takes as input a language embedding $\vec{l}_e$ from a given epoch of training, $e$, and outputs the typological class a language belongs to (as coded in WALS).
When $e$ is 0, this indicates the pre-trained language embeddings as obtained from \citet{ostling_tiedemann:2017}.
Increasing $e$ indicates the number of epochs of PoS tagging during which the language embedding has been updated.
All results are the mean of three-fold cross-validation.
We are mainly interested in observing two things: i) Which typological traits do language embeddings encode?; ii) To what extent can we track the changes in these language embeddings over the course of fine-tuning for the task of PoS tagging?.

\begin{figure*}[htbp]
	\centering
	\includegraphics[width=0.7\textwidth]{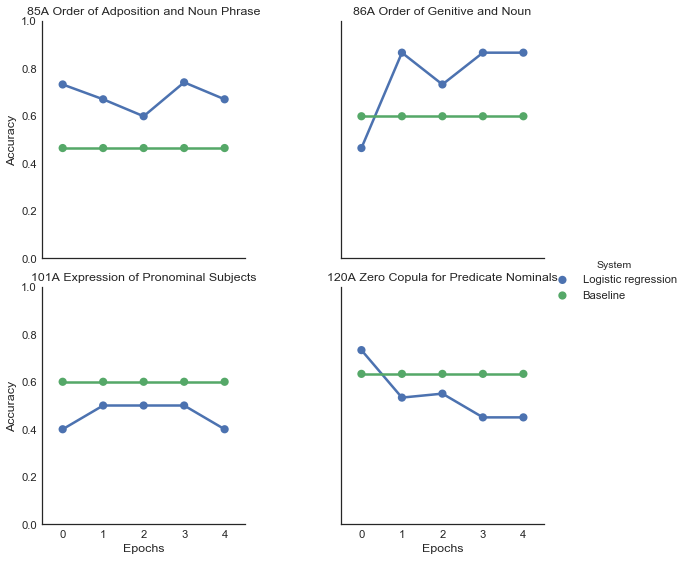}
    \caption{\label{fig:typology}Predicting typological features in WALS. The x-axes denote number of epochs the language embeddings have been fine-tuned for. The y-axe denotes classification accuracy for the typological feature at hand.}
\end{figure*}

We train the neural network model over five epochs, and investigate differences of classification accuracies of typological properties as compared to pre-trained embeddings.
A baseline reference is also included, which is defined as the most frequently occurring typological trait within each category.
In these experiments, we disregard typological categories which are rare in the observed sample (i.e. of which we have one or zero examples).
Looking at classification accuracy of WALS features, we can see four emerging patterns:
\begin{enumerate}
    \setlength\itemsep{-3pt}
	\item The feature is pre-encoded;
    \item The feature is encoded by fine-tuning;
    \item The feature is not pre-encoded;
    \item The feature encoding is lost by fine-tuning.
\end{enumerate}
One example per category is given in Figure~\ref{fig:typology}.
Two features based on word-ordering can be seen as belonging in the categories of features which are either pre-encoded or which become encoded during training.
The fine-tuned embeddings do not encode the feature for whether pronominal subjects are expressed, or the feature for whether a predicate nominal has a zero copula.


\subsubsection{Predicting Uralic typological features}

\begin{figure*}[htbp]
	\centering
	\includegraphics[width=0.7\textwidth]{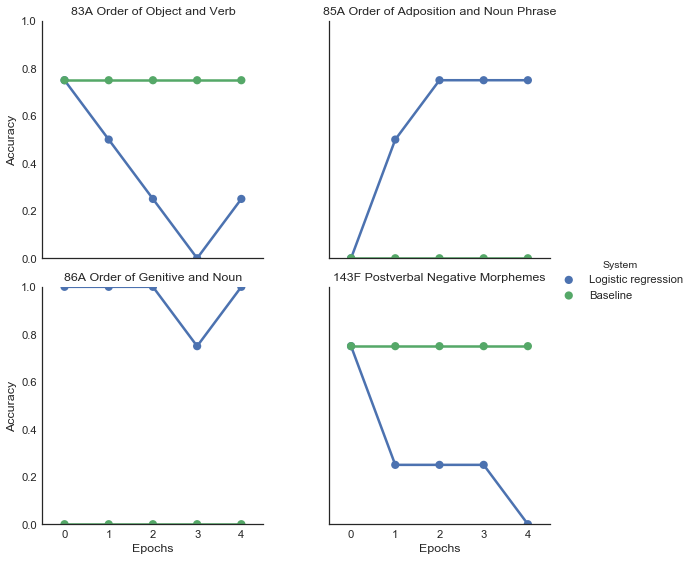}
    \caption{\label{fig:uralic_typology}Predicting typological features in Uralic languages. The x-axes denote number of epochs the language embeddings have been fine-tuned for. The y-axes denote classification accuracy for the typological feature at hand.}
\end{figure*}

Finally, we attempt to predict typological features for the four Uralic languages included in our sample, as shown in Figure~\ref{fig:uralic_typology}.
Similarly to the larger language sample in Figure~\ref{fig:typology}, the Uralic language embeddings also both gain typological information in some respects, and lose information in other respects.
For instance, the pre-trained embeddings are not able to predict ordering of adpositions and noun phrase in the Uralic languages, whereas training on PoS tagging for two epochs adds this information.

\section{Discussion}

\subsection{Language embeddings for Uralic model transfer}
In the monolingual transfer setting, we observed that transferring from more closely-related languages was relatively beneficial.
This is expected, as the more similar two languages are, the easier it ought to be for the model to directly apply what it learns from one language to the other.
Concretely, we observed that transferring between the two Finnic languages in our sample, Finnish and Estonian, worked relatively well.
We further observed that including language embeddings in this setting had little-to-no effect on the results.
This can be explained by the fact that the language embedding used is the same throughout the training phase, as only one language is used, hence the network likely uses this embedding to a very low extent.

In bilingual settings, omitting the language embeddings results in a severe drop in tagging accuracy in most cases.
This is likely because that treating our sample of languages as being the same language introduces a large amount of confusion into the model.
This is further corroborated by the fact that treating the two Finnic languages in this manner results in a relatively small drop in accuracy.

Including  language embeddings allows for the model transfer setting to be beneficial for the more closely related languages.
This bodes well for the low-resource case of many Uralic languages in particular, and possibly for low-resource NLP in general.
In the cases of the more distantly related language pairings, including language embeddings does not result in any significant drop in accuracy.
This indicates that using language embeddings at least allows for learning a more compact model without any significant losses to performance.

\subsection{Language embeddings for Uralic typology}
Interestingly, the language embeddings are not only a manner for the neural network to identify which language it is dealing with, but are also used to encode language similarities and typological features.
To contrast, the neural network could have learned something akin to a one-hot encoding of each language, in which case the languages could easily have been told apart, but classification of typological features would have been constantly at baseline level.

Another interesting finding is the fact that we can track the typological traits in the distributed language representations as they are fine-tuned for the task at hand.
This has the potential to yield insight on two levels, of interest both to the more engineering-oriented NLP community, as well as the more linguistically oriented CL community.
A more in-depth analysis of these embeddings can both show what a neural network is learning to model, in particular.
Additionally, these embeddings can be used to glean novel insights and answer typological research questions for languages which, e.g., do not have certain features encoded in WALS.

In the specific case of Uralic languages, as considered in this paper, the typological insights we gained are, necessarily, ones that are already known for these languages.
This is due to the fact that we simply evaluated our method on the features present for the Uralic languages in WALS.
It is nonetheless encouraging for this line of research that we, e.g., could predict WALS feature 86A (\textit{Order of Genitive and Noun}) based solely on these embeddings, and training a very simple classifier on a sample consisting exclusively of non-Uralic languages.

\section{Conclusions and future work}
We investigated model transfer between the four Uralic languages Finnish, Estonian, Hungarian and North Sámi, in PoS tagging, focussing on the effects of using language embeddings.
We found that model transfer is successful between these languages, with the main benefits found between the two Finnic languages (Finnish and Estonian), when using language embeddings.
We then turned to an investigation of the typological features encoded in the language embeddings, and found that certain features are encoded.
Furthermore, we found that the typological features encoded change when fine-tuning the embeddings.
In future work, we will look more closely at how the encoding of typological traits in distributed language representations changes depending on the task on which they are trained.

\section*{Acknowledgements}
The authors would like to thank Iina Alho for help with translating the abstract to Finnish.
We would also like to thank Robert Östling for giving us access to pre-trained language embeddings.

\bibliographystyle{acl_natbib}
\bibliography{naaclhlt2018}

\end{document}